\documentclass[sigconf,nonacm]{acmart} 

\usepackage{balance}

\title[Agentic LLM Systems]{Practical Considerations for Agentic LLM Systems}

\author{Chris Sypherd}
\affiliation{
  \institution{University of Edinburgh}
  \city{Edinburgh}
  \country{United Kingdom}}
\email{c.n.sypherd@sms.ed.ac.uk}

\author{Vaishak Belle}
\affiliation{
  \institution{University of Edinburgh}
  \city{Edinburgh}
  \country{United Kingdom}}
\email{vbelle@ed.ac.uk}

\begin{abstract}
As the strength of Large Language Models (LLMs) has grown over recent years, so too has interest in their use as the underlying models for autonomous agents. Although LLMs demonstrate emergent abilities and broad expertise across natural language domains, their inherent unpredictability makes the implementation of LLM agents challenging, resulting in a gap between related research and the real-world implementation of such systems. To bridge this gap, this paper frames actionable insights and considerations from the research community in the context of established application paradigms to enable the construction and facilitate the informed deployment of robust LLM agents. Namely, we position relevant research findings into four broad categories—Planning, Memory, Tools, and Control Flow—based on common practices in application-focused literature and highlight practical considerations to make when designing agentic LLMs for real-world applications, such as handling stochasticity and managing resources efficiently. While we do not conduct empirical evaluations, we do provide the necessary background for discussing critical aspects of agentic LLM designs, both in academia and industry.
\end{abstract}

         
\newcommand{\BibTeX}{\rm B\kern-.05em{\sc i\kern-.025em b}\kern-.08em\TeX}

\begin{document}

\pagestyle{fancy}
\fancyhead{}

\maketitle 

\section{Introduction}

In academia, the concept of "agents" has been well-defined for decades (e.g.,~\cite{Wooldridge_Jennings_1995}), and thus the proposition of agents based on LLMs comes with predefined criteria and expectations. As such, agentic LLMs in the research community have come to be defined as autonomous systems with capabilities of beliefs~\cite{sclar-etal-2023-minding, li-etal-2023-theory, GenAgentsSimulcraPark2023}, reasoning~\cite{Zhang2024LLMAA}, planning~\cite{huang2024understandingplanningllmagents, shen2023hugginggpt}, and control~\cite{shen2023hugginggpt}. Under this definition, the ability to plan, reason, and interact with an environment have emerged as the key considerations for success~\cite{xi2023risepotentiallargelanguage}.

\begin{figure}[h]
  \centering
  \includegraphics[width=0.85\linewidth]{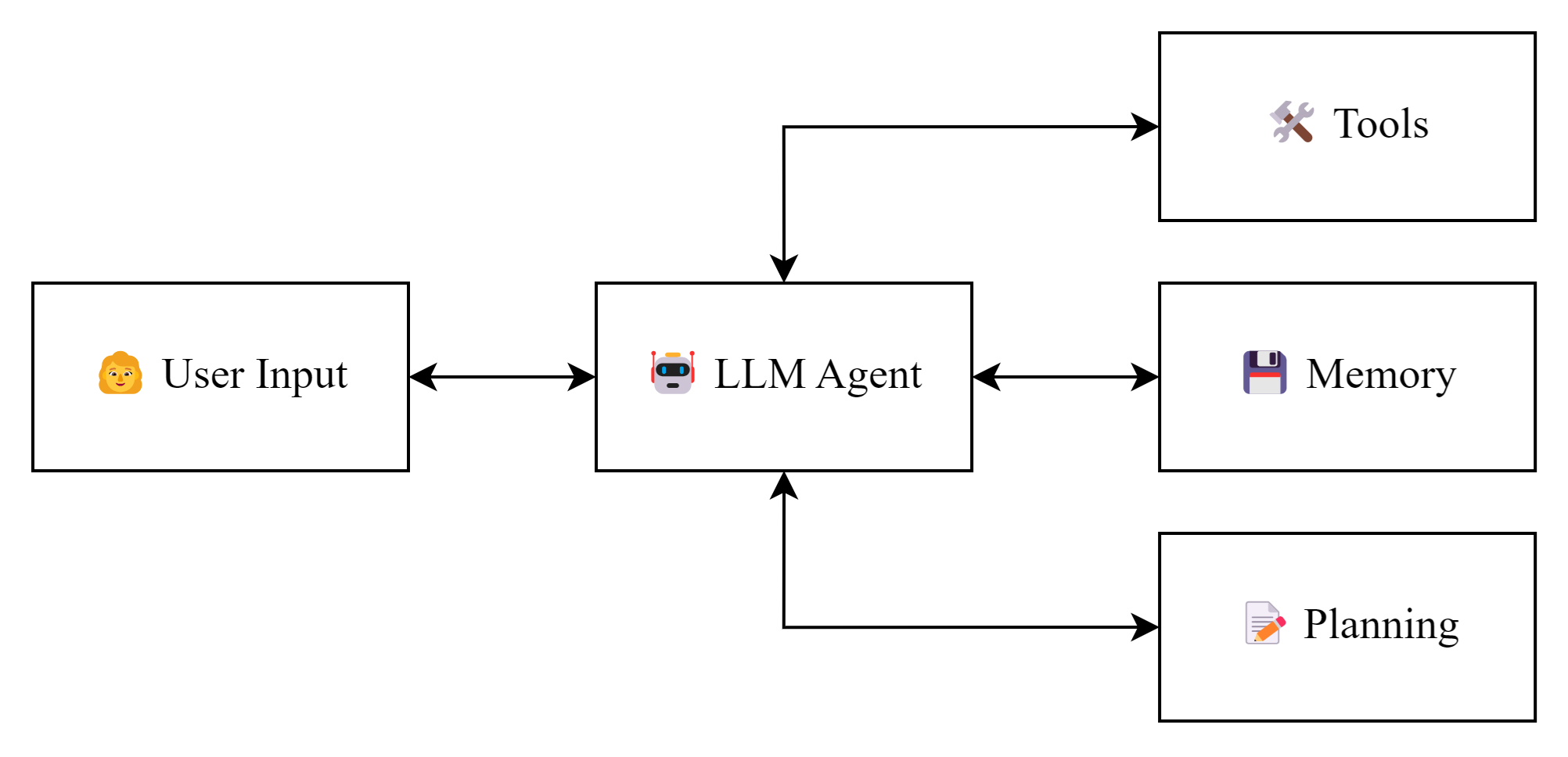}
  \caption{A typical application-focused depiction of LLM agents.}
  \label{fig:IndustryAgent}
  \Description{A diagram showing five boxes containing the texts "User Input," "Agent," "Planning," "Tools," and "Memory. The "Agent" box sits between "User Input" and the other three boxes.}
\end{figure}

For LLM agents in industry and real-world deployment, the history and breadth of agents has been condensed to a definition along the lines of "a system that can use an LLM to reason through a problem, create a plan to solve the problem, and execute the plan with the help of a set of tools"~\cite{NvidiaAgent}. Most industry discussions follow this form, introducing the LLM as the central reasoning engine and adding planning, memory, and tools as three necessary modules (e.g., ~\cite{SAAgent, PGAgent, TFAgent, NvidiaAgent, LWAgent}). Indeed, most industry resources focusing on deployable agentic LLM systems are accompanied by a diagram similar to Figure \ref{fig:IndustryAgent}, focusing largely on single agents. While this description is helpful for the most basic of LLM agents, it glosses over some of the more nuanced considerations that must be made for the informed construction of robust agentic LLM systems.

The prevailing view of LLM agents in industry brings to light the disparity between (1) research into LLMs and agents and (2) the application of agentic LLM systems in real-world scenarios. To bridge this gap, we propose framing relevant findings from the research community in the common industry view of LLM agents. To that end, we organize this work into four main sections—Planning, Memory, Tools, and Control Flow—that correspond to, respectively, planning, memory, tools, and the central reasoning engine components referenced above. We tailor the contents of this paper to black-box LLM-based single-agent systems typical of that industry perspective. By doing so, we hope to create an actionable and approachable survey that enables information exchange between academia and industry within a manageable scope.

\section{Related Work}

Many surveys discussing LLM-based agents focus on multi-agent frameworks and related ideas~\cite{Guo2024LargeLM, Zhang2024LLMAA}. While similar, the challenges facing multi-agent systems are distinct from the real-world deployment of a single LLM agent. Here, we focus more on deliberately crafting a robust agent rather than the orchestration of many.

Another approach, taken by~\cite{yang2024llmwizardcodewand}, focuses on methods for improving agentic LLM performance starting at the underlying model, looking into data composition and training methodologies. We focus on implementation considerations that improve agentic LLM system performance from a black-box perspective, which lends itself more to real-world deployment. Others focus on creating unified taxonomies ~\cite{li2024reviewprominentparadigmsllmbased, xi2023risepotentiallargelanguage} or target a single component of LLM agents, such as planning~\cite{huang2024understandingplanningllmagents}. 

The most similar work is~\cite{Wang_2024}, providing a comprehensive survey of works relating to LLMs as agents as well as reviewing aspects of their design, application, and evaluation. While~\cite{Wang_2024} develops a valuable unified framework based on extant research, we leverage research findings to provide practical application-focused insights and frame our review in the context of the LLM agent paradigm that has developed organically in industry.

To the best of our knowledge, this is the first work that coalesces research relevant to LLM agents through the lens of common industry practices. We expand that contribution by not just presenting existing research but by extrapolating actionable insights and best practices from it.

\section{Applied Scenario}

\begin{figure}[h]
  \centering
  \includegraphics[width=0.85\linewidth]{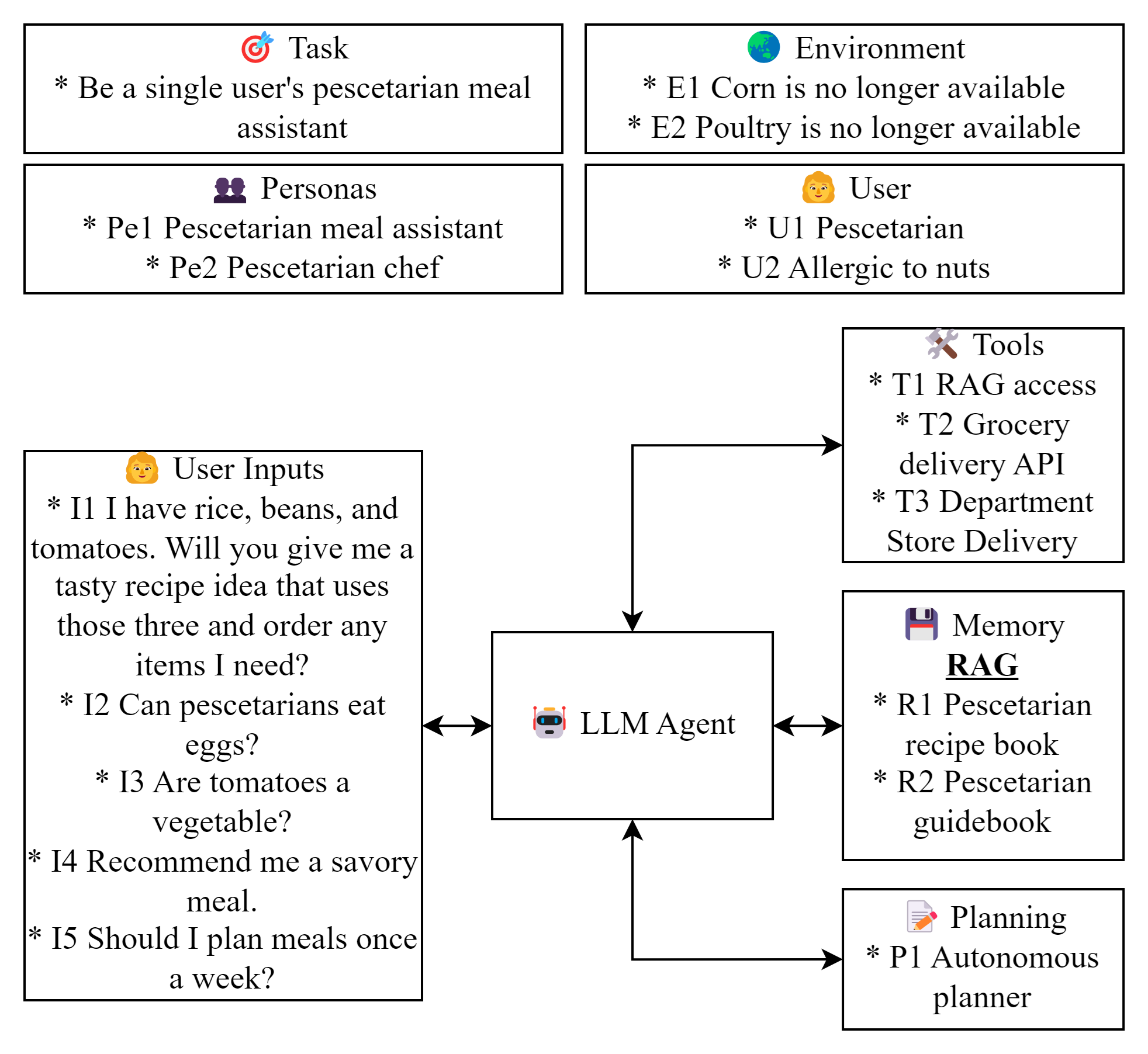}
  \caption{An example scenario featuring a pescetarian meal assistant LLM agent.}
  \label{fig:Example}
  \Description{A diagram showing the user, task, and environment definition relating to a pescetarian meal assistant example.}
\end{figure}

To help illustrate some of the following points, we propose the example outlined in Figure \ref{fig:Example} of applying an LLM agent as a pescetarian\footnote{Someone that does not eat meat, aside from fish and other seafood.} meal assistant. We will refer to this as the primary example\footnote{While we attempt to select a simple example with some relevance to the real world, it is still a contrived example to demonstrate the points outlined in this work and may not fully reflect the complexities of the real world.} throughout this work for consistency, citing specifics from it by the codes assigned in Figure \ref{fig:Example} (e.g., \ref{fig:Example}.R1 to refer to "Pescetarian recipe book").

\section{Glossary}

This glossary serves to briefly introduce the following terms that will be used across subsequent sections. Later sections will provide additional contextualization and examples of their utility but not necessarily explicit definitions.

\textbf{Persona.} A persona (also referred to as a "role" or a "profile") is the identity assigned to the LLM, often as part of the system prompt. The persona is the lens through which the LLM will interpret and respond to prompts. The persona (e.g., Figure \ref{fig:Example}.Pe2) can be defined and refined by an occupation (e.g., "professional chef"), level and domain of expertise (e.g., "specializing in pescetarian dishes"), and personality traits (e.g., "friendly and understanding") but can be further customized by adding details such as age, race, gender, and nationality~\cite{Argyle_2023, Wang_2024}. 

\textbf{Tool.} \label{tool} Tools are the means by which an LLM can interact with its environment (beyond basic textual exchange) and access external resources~\cite{qin2024toollearningfoundationmodels}. \nameref{rag} (RAG) is commonly used as a tool (Figure \ref{fig:Example}.T1) but is limited in its utility as it exclusively \textit{retrieves} information about the environment. The true power of tools is realized when they are used to perform actions in the environment, such as the example in Figure \ref{fig:Example}.T2 that would allow the pescetarian meal assistant to not only recommend recipes but also order the ingredients to prepare them. Other examples of tools include ground-truth verification methods (e.g., code execution and calculator usage) and real-time environment querying (e.g., requesting trending recipes)~\cite{wang2024what}.

\textbf{Hyperparameters.} \label{hyperparameters} This section includes a brief overview of hyperparameters we will reference but does not explore their technical details\footnote{See~\cite{GeminiDocs, OpenAIDocs, AnthropicDocs} for common commercial API support for these hyperparameters.}.

\begin{itemize}
    \item \textbf{Seed.} Some LLM interfaces will have a “seed” parameter that should, provided all other parameters remain constant, produce the same output.
    \item \textbf{Temperature.} Temperature corresponds to the degree to which randomness will be employed in selecting the output tokens. This usually plays out in higher temperature responses being more creative and rambling while lower temperature responses are more predictable and straight-to-the-point. Thus, for more consistent results, a lower temperature (e.g., 0.0 to 0.5) can be used.
    \item \textbf{Top-p.} Top-p (also known as "nucleus sampling;" introduced in~\cite{Holtzman2020The}) corresponds to the probability threshold for selecting tokens that can form part of the output, bounded 0.0 to 1.0. Lower top-p values restrict the pool of tokens the LLM can choose from, resulting in more reproducible outputs.
\end{itemize}

\section{Planning} \label{planning}

Planning has long been a core component of agent research~\cite{AutoPlan, Wooldridge_Jennings_1995}; it allows more complex tasks to be handled in smaller, more manageable steps. Planning can also enhance the interpretability of an LLM agent, as the steps of the plan and the stopping criteria will be defined in a interpretable format.

\subsection{LLMs and Planning}

Despite anecdotal applications showing signs of successful LLM planning~\cite{CVLLMPlannerSong}, more holistic reviews suggest that LLMs make poor planners~\cite{valmeekam2022large, ValmeekamPlanning2024, Kambhampati_2024, liu2023llmpempoweringlargelanguage, Dagan2023DynamicPW}. As such, if an LLM agent is to be deployed in an environment with a consistent task, manually curating a plan can alleviate the pains of poor LLM planning as well as provide an opportunity to manually craft relevant roles and prompts. Another option is to augment the LLM agent with an external planning tool, which has shown promise~\cite{liu2023llmpempoweringlargelanguage, Dagan2023DynamicPW}. Because LLM planning remains an open area of research, the example in Figure \ref{fig:Example}.P1 simply assumes planning capabilities without subscribing to a specific approach, for illustrative purposes.

To describe current approaches to planning in LLM agents, we categorize them into \textit{implicit} and \textit{explicit} planning. For \textbf{implicit planning}, some agents will rely on the LLM to iteratively determine the immediate next step until the task is complete, without ever eliciting a plan~\cite{zhang-etal-2024-pddlego, gur2024a}. This approach relies on the idea that, when provided with an end goal, the LLM can maintain an internal plan whose steps are revealed iteratively without any explicit plan formalization. This approach can be viable when the environment is dynamic or only partially observable, such as interacting with a webpage~\cite{gur2024a}. The other form of implicit planning is the creation and execution of a plan in a single inference, as seen in prompting strategies such as Plan-and-Solve~\cite{wang-etal-2023-plan} and, to a degree, zero-shot Chain-of-Thought~\cite{kojima2022large}. These approaches rely on conditioning subsequent token generation (i.e., the "execution") on a plan by first generating said plan. Due to the single-hop nature of this approach, it is not recommended for complex tasks, particularly those that would benefit from feedback during execution. 

\textbf{Explicit planning} is characterized by the explicit formalization of a multi-step plan, typically executed in a multi-hop fashion. The most basic form is to simply request the formulation of a plan and then execute it, as demonstrated by the Least-to-Most prompting strategy~\cite{zhou2023leasttomost}. More advanced approaches will first develop a plan and then iteratively refine the plan as steps are executed~\cite{liu2024reasonfutureactnow}. Both of these require long-term planning, which is where LLMs tend to demonstrate lackluster performance~\cite{valmeekam2022large, ValmeekamPlanning2024, Kambhampati_2024}.

\subsection{Task Decomposition} \label{task-decomp}

It is important to understand the limitations of an LLM before formulating a plan for it to execute. Agentic LLM systems are often applied to problems that a single LLM call cannot resolve but a sequence of calls can. Tasks can typically be decomposed into smaller pieces that, when solved individually, can be reconstructed to produce the final solution~\cite{zhou2023leasttomost}.

Returning to Figure \ref{fig:Example}, the request made in Figure \ref{fig:Example}.I1 is composed of multiple subtasks, namely: (1) retrieve recipes that contain rice, beans, and tomato that the user will like and (2) order any missing ingredients. It is also reasonable to decompose (1) further, into a retrieval of recipes that contain the required ingredients and separately a request to select the one that best fits the user's tastes.

If decomposing a well-defined task manually, iteratively decomposing the task into subtasks and testing an LLM on them can provide valuable insight into what the LLM can consistently handle. Breaking down the problem logically is simple enough, but ascertaining which tasks an LLM can perform well and which require further decomposition can be challenging, particularly when dealing with stochasticity and prompt changes. It is recommended to evaluate the LLM agent frequently and systematically during this process, as discussed in Section \ref{eval}. It may be easier to start at the most basic building blocks of the tasks and combine them than to find the minimum number of viable tasks to start.

While it may be intuitive to assume that the more atomic the task the better, this is not always the case. It has been shown that LLMs not only possess the ability to solve multiple distinct tasks in a single query~\cite{xiong2024oncellmsincontextlearn, son-etal-2024-multi-task, laskar-etal-2023-systematic} but that composing multiple tasks into a single prompt can \textit{increase} performance on all constituent tasks, as well as decreasing overall context usage~\cite{son-etal-2024-multi-task}. However, the degree to which tasks may be combined should be the subject of rigorous experimentation for the specific task and environment in which it is considered.

\subsection{Plan Adherence} \label{p-adherence}

One of the responsibilities of the LLM agent is to oversee the application of the plan. It should decide if a step needs to be repeated (e.g., for \nameref{error-handling}) or skipped for a given input (e.g., to iterate on the plan~\cite{liu2024reasonfutureactnow}). One of the major concerns of LLMs as planners is their inability to identify whether or not they can complete a given task~\cite{Kambhampati_2024}. As such, it is often impossible for an LLM agent to know if a step will be successful until it has been attempted. Thus, it follows logically that an evaluation of the success of each step should take place following execution. Similarly, the overall success of the plan should be evaluated upon completion of all steps. If unsuccessful, the LLM agent may need to adjust or rerun the plan, based on the results of each step and the overall plan (see Section \ref{error-handling} for a discussion on incorporating feedback).

\section{Memory} \label{memory}

\subsection{Retrieval Augmented Generation} \label{rag}

Retrieval augmented generation (RAG) (introduced in~\cite{lewis2021retrievalaugmentedgenerationknowledgeintensivenlp}) has emerged as a staple of agentic LLM applications in industry ~\cite{gao2024retrievalaugmentedgenerationlargelanguage}. The basis is simple: a system that can provide external context relevant to a natural language input. Typically, an incoming input will be compared against a ground-truth data store and the most relevant piece(s) of information will be provided to the LLM as context upon which it will base its response. This can be done either implicitly, where a user's input is always used for retrieval for a given LLM call, or explicitly, where the LLM uses RAG as a tool. This has a number of benefits for LLM systems:

\begin{itemize}
  \item \textbf{Grounding.}  Rather than relying on the LLM “remembering” relevant context from its training data correctly, we can provide the LLM with accurate relevant information. Providing grounded text as context significantly reduces LLM hallucinations and fills knowledge gaps in the training data~\cite{shuster2021retrievalaugmentationreduceshallucination, es-etal-2024-ragas, lewis2021retrievalaugmentedgenerationknowledgeintensivenlp}.
  \item \textbf{Explainability.} Rather than relying on an LLM opaquely referencing information it has been trained on, adherence to context supplied as part of RAG provides insight into exactly where an LLM is getting its information~\cite{gao2024retrievalaugmentedgenerationlargelanguage}.
  \item \textbf{Timeliness.} While LLMs can reference information from their static training data, the LLM will be subject to a hard information cutoff (the latest date training data was scraped) and a soft information cutoff (events close to its hard information cutoff that have limited coverage). Rather than turning to the infeasible prospect of retraining with updated data, we can \textit{provide} updated information that is relevant to the query as context~\cite{gao2024retrievalaugmentedgenerationlargelanguage}.
  \item \textbf{Outsourcing.} Depending on the content, quality, and reliability of the RAG database, aspects of the query can be implicitly outsourced to the context returned, such as reasoning and decision-making.
  \item \textbf{Alignment.} The vast amount of training data used for LLMs is the source of their natural language understanding but should not necessarily be relied on for unbiased, trustworthy, and safe generation. Typically, aligning LLM ouputs with human preferences is seen as a data collection and training problem~\cite{wang2023aligninglargelanguagemodels, bai2022constitutionalaiharmlessnessai} but can also be addressed post-hoc with RAG. By augmenting an LLM’s natural language capabilities and tendencies with context derived from a more refined dataset that adheres to a desired set of human preferences, its output can be guided to conform to a desired set of content and attitudes. This requires careful curation of the data store but is a viable method for black-box alignment.
\end{itemize}

To exemplify the points above, consider the RAG sources referenced in Figure \ref{fig:Example}.R1 and \ref{fig:Example}.R2. Figure \ref{fig:Example}.R1 can be quoted to avoid recipe hallucinations (grounding) and be updated with new recipes (timeliness). Figure \ref{fig:Example}.R2 could be useful in responding to Figure \ref{fig:Example}.I2; where there is no general consensus, we can supply our own ground truth rather than require the LLM to answer a potentially moral question (outsourcing). The tone and terminology of both Figure \ref{fig:Example}.R1 and \ref{fig:Example}.R2 will guide the ideals, content, and terminology used by the LLM (alignment).

There are two main approaches to RAG: knowledge graphs~\cite{edge2024localglobalgraphrag} and vector databases~\cite{gao2024retrievalaugmentedgenerationlargelanguage, barron2024domainspecificretrievalaugmentedgenerationusing}, with the latter seeing far greater adoption due to its simplicity. For a discussion on implementing RAG and the extant commercial and open-source offerings, see~\cite{gao2024retrievalaugmentedgenerationlargelanguage}.

\subsection{Long-Term Memory}

Sometimes, key information is gained during a conversation that may be helpful across all contexts, such as a useful piece of external knowledge or information about a user or task. In those instances, it may be advantageous to store that information in a way accessible to the agent so that its impact is not limited to the current context. This is commonly referred to as "long-term memory"~\cite{qian-etal-2024-chatdev, Wang_2024, Zhong2023MemoryBankEL}\footnote{A real-world implementation of long-term memory is OpenAI’s “Memory”~\cite{OAIMemory}. During conversations with ChatGPT, the LLM will save information that it deems particularly useful to its memory. That memory is then made available to the LLM in future conversations.  A clear benefit of this is that it reduces repetition on the part of the user and allows the LLM to better fulfill its objective of providing relevant responses.}.

We want to be selective with the information that is stored in long-term memory so that it is generally useful and not excessively large. Some common approaches are to store prior solutions to queries~\cite{qian-etal-2024-chatdev}, global summaries and insights~\cite{Zhong2023MemoryBankEL}, and acquired tools~\cite{wang2023voyageropenendedembodiedagent}.

Long-term memory can be enhanced with reflection, consolidation, forgetting, revision, and other mechanisms designed to mimic long-term memory in humans (see~\cite{Zhong2023MemoryBankEL} for a discussion on advanced long-term memory implementation). For simplicity, we focus on a simple version of long-term memory, where information is simply stored and retrieved, and any edits are manual. For this simple variation, we derive the following three criteria from existing literature on long-term memory in LLM agents~\cite{qian-etal-2024-chatdev, Wang_2024, Zhong2023MemoryBankEL, OAIMemory, wang2023voyageropenendedembodiedagent} to use as a litmus test for what information should be stored:

\begin{itemize}
    \item \textbf{Independent.} The information should not have any implicit dependencies, such as input values.
    \item \textbf{Relevant to a consistency.} The information should be relevant to \textit{consistencies} in the agentic LLM system, which may include a task, user, or environment. 
    \item \textbf{Applicable long-term.} The information should consistently be applicable to contexts to which the LLM agent may be exposed.
\end{itemize}

See Table \ref{tab:in-context-criteria} for the above criteria applied to examples drawn from Figure \ref{fig:Example}.

If the three criteria above are ensured, then the gathered in-context information can be a useful starting place for prompt improvements. It will be information that has been identified as generally and consistently useful to the LLM agent's environment and may be appropriately suited to permanent inclusion in user or system prompts. It can also be valuable to review long-term memory when making prompt, task, persona, hyperparameter, or model updates; reordering LLM calls; or adjusting tool functionality as such changes may impact the validity of the three criteria.

\begin{table}[t]
	\caption{Analysis of information from Figure \ref{fig:Example} in context of the three criteria for storing in-context information.}
	\label{tab:in-context-criteria}
	\begin{tabular}{p{58pt}|p{40pt}|p{35pt}|p{39pt}|p{20pt}}\toprule
		\textit{Information} & \textit{Independent} & \textit{Relevant to Consistency} & \textit{Applicable Long-Term} & \textit{Store the Info} \\ \midrule
		E1 Corn is no longer available & Yes & Yes & Yes & Yes \\ E2 Poultry is no longer available & Yes & No & No & No \\ U2 Allergic to nuts & Yes & Yes & Yes & Yes \\ I1 I have rice, beans, and tomatoes... & No & Yes & No & No \\ \bottomrule
	\end{tabular}
\end{table}

\subsubsection{Extracting Information for Long-Term Memory}

A common approach to extracting information that belongs in long-term memory is to leverage an external conversation moderator~\cite{Zhong2023MemoryBankEL, Reflexion}. The external moderator (e.g., an LLM with a separate role) that reviews conversations (either whole or in pieces) and can be tasked with extracting information it deems compliant with the three criteria above. This is an instance where care must be taken with phrasing as the subjectivity of the task may make the LLM prone to framing bias in its response (e.g., if we ask if there is anything useful to pull out, the LLM will likely pull out some information)~\cite{Echterhoff2024CognitiveBI}.

\subsubsection{Storing Long-Term Memory}

Once a piece of information has been deemed worthy of long-term memory, it should be stored. Some approaches include embedding and storing the information in a vector database (similar to \hyperref[rag]{RAG})~\cite{Zhong2023MemoryBankEL, lin2023agentsimsopensourcesandboxlarge} and natural language storage, although interacting with the latter quickly becomes unwieldy as the amount of long-term memory increases. The structure of the vector database allows us to easily query relevant information~\cite{lin2023agentsimsopensourcesandboxlarge, wang2023voyageropenendedembodiedagent, Zhong2023MemoryBankEL}.

\subsubsection{Utilizing Long-Term Memory}

Once information is stored in long-term memory, we must decide when to expose it to the LLM agent. It is key to understand what information is relevant in the current scope. LLM agents may be composed of many LLM calls with different purposes and contexts; not all information from long-term memory will apply to every LLM call. For example, if the user from Figure \ref{fig:Example} decides to plan meals once a week (per Figure \ref{fig:Example}.I5), that would be a valuable long-term memory for Figure \ref{fig:Example}.Pe1 but not necessarily \ref{fig:Example}.Pe2, which is mainly used for its culinary expertise. In such instances, the relevance afforded by retrieval from a vector database is valuable~\cite{lin2023agentsimsopensourcesandboxlarge, wang2023voyageropenendedembodiedagent, Zhong2023MemoryBankEL}. Once relevant information has been retrieved from long-term memory, it can be shared in an LLM call via the user or system prompt.

\section{Tools} \label{tools}

\subsection{Using Tools} \label{using-tools}

To enable the LLM to use tools, tool descriptions and methods of invocation need to be exposed with the LLM (similar to traditional software engineering documentation). If the number of tools in use is small, they can be introduced in natural language. The method for invoking a tool should be clear and easily parsable. A common way to do this is by defining JSON schemas or function signatures, although the latter has been shown to be better for LLM agents~\cite{BeatingGaia, wang2024executable}.

Tools can be called either explicitly or implicitly, with the former being the de facto approach in practice. Explicit usage simply entails the invocation of a tool as part of the LLM agent's output~\cite{schick2023toolformer, qin2024toolllm}. Once the tools are defined and passed as context, the agent will have the means to perform such an invocation in the specified parsable format. Tools can also be implicitly invoked by the implementor in response to an LLM agent's action or inaction. For example, if a transition between personas occurs, it may be the case that the system will always benefit from a summarization of preceding dialogue. Rather than rely on the LLM agent to invoke a summarization at every persona change, every such transition can trigger a summarization behind the scenes. See Section \ref{trad-eng} for a discussion on incorporating implicit tool calling.

\subsection{Managing Multiplicity}

As the number of tools grows, defining tools in natural language quickly becomes unwieldy and a structured approach is necessary. To do so, we can leverage LLMs' convenient understanding of code by creating more concise tool definitions using JSON schemas or function signatures in conjunction with condensed natural language descriptions\footnote{See https://python.langchain.com/docs/concepts/\#tools for a discussion on tool definition and \cite{PConeTools} for an implementation.}.

Often, distinct tools can be placed into distinct groups based on similar core functionality (i.e., if they can reasonably be seen as inheriting from the same base class). These groups can be called “toolsets” or "toolkits"\footnote{See https://python.langchain.com/docs/concepts/\#toolkits.} and are helpful for determining if tools can be combined behind a single interface or introduced together in the prompt. For example, the tools Figure \ref{fig:Example}.T1 and \ref{fig:Example}.T2 introduced in the example would not belong in the same toolset but \ref{fig:Example}.T2 and \ref{fig:Example}.T3 would.

\subsection{Adding Tools Dynamically}

Sometimes the tools that are available in the environment in which an agentic LLM system will be deployed are not known beforehand. In this case, we can add “tool identification” as a task for the system~\cite{schick2023toolformer, wang2023voyageropenendedembodiedagent}. A compelling example and implementation of this can be observed in the Voyager paper, where an LLM-based agent autonomously traverses the world of Minecraft\footnote{https://www.minecraft.net} and dynamically assembles a set of tools based on interactions with the environment, which are then stored in long-term memory~\cite{wang2023voyageropenendedembodiedagent}.

\section{Control Flow} \label{control-flow}

In the context of LLM agents, control flow refers to the ability to determine what needs to be done in order to respond to a query. Tasking an LLM with control flow is what enables LLM-based agents to accomplish complex tasks that elude the capacity of a single inference. This endows the LLM agent with the autonomy to incorporate advanced techniques such as planning, tool usage, and multi-step reasoning as it sees fit~\cite{shen2023hugginggpt, Wang_2024}.

In practice, this may look like the LLM agent receiving user input (i.e., observing the environment) and selecting the immediate next action. The agent continues to take actions until it decides to stop. For this to be possible, the LLM agent needs to be aware of the action space~\cite{yao2023react}, such as the stopping criteria, available tools (e.g., Figure \ref{fig:Example}.T1, \ref{fig:Example}.T2, and \ref{fig:Example}.T3), available planning options (\ref{fig:Example}.P1), the ability to take a turn to think out loud~\cite{yao2023react}, and utilizing other personas (e.g., \ref{fig:Example}.Pe2).

Consider an LLM agent receiving Figure \ref{fig:Example}.I1. Rather than simply providing an output, the agent can opt to leverage \ref{fig:Example}.P1, the planning module, to decompose the complex task and generate a multi-step plan that it can then administer. Once the plan is complete, the agent can decide if it has enough information to provide the final output to \ref{fig:Example}.I1 or if it needs to take additional actions.

Here, we present practical considerations for ensuring the LLM agent can interact with its environment smoothly and without interruption.

\subsection{Output Processing} \label{io-processing}

When chaining together multiple LLM inputs and outputs, it is often advantageous to process the text before handing it off to the next step. Although natural language is human-readable, it is advisable to use a more structured format (such as JSON or executable code) that is easily parsable~\cite{wang2024executable}. While weaker models may struggle with instruction following, most commercial models have been optimized to adhere to desired output formats specified in the user or system prompt\footnote{Output processing documentation for common commercial models: Anthropic~\cite{AnthropicOutput}; OpenAI~\cite{OAIOutput}; Google~\cite{GeminiOutput}}.

Because we approach LLMs from a black-box perspective, we do not discuss the underlying approaches to constraining LLMs to output a specific format. However, it is important to note that the reasoning capabilities demonstrated by an LLM may be negatively (and inadvertently) impacted by constrained generation, depending on the implementation~\cite{BeurerKellner2024GuidingLT, tam2024letspeakfreelystudy}. Because of this, it has been shown that requiring code outputs instead of a specific structure can yield better agents~\cite{wang2024executable}.

\subsection{Error Handling} \label{error-handling}

Error handling is one of the most important yet elusive parts of building a robust agentic LLM system. Because LLMs are inherently stochastic, chaining several LLM calls together compounds the risk of failure to the point of near inevitability for long sequences. As such, every LLM call in an agentic LLM system should be treated as a potential point of failure and supported by appropriate error handling.

\begin{figure}[h]
  \centering
  \includegraphics[width=0.85\linewidth]{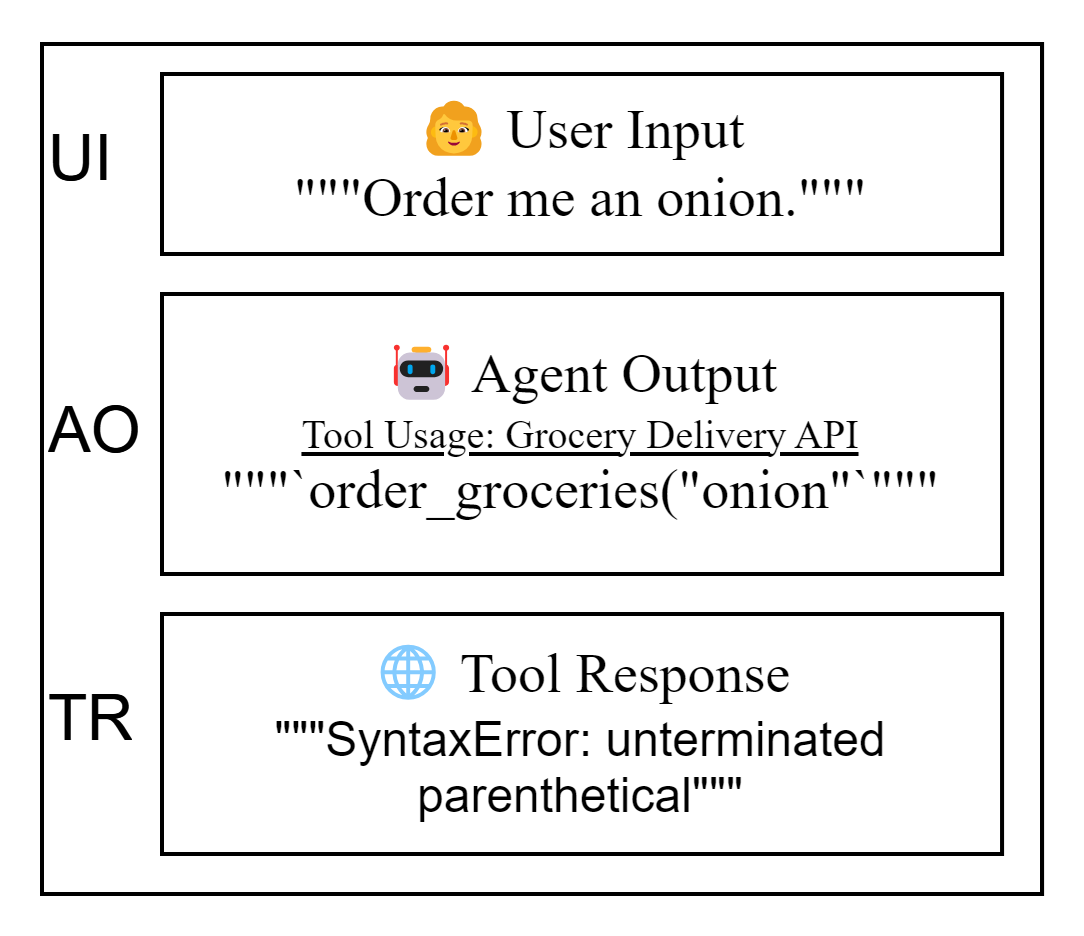}
  \caption{An example of an erroneous tool call, following the scenario presented in Figure \ref{fig:Example}.}
  \label{fig:Error}
  \Description{A diagram showing an erroneous tool call from the agent.}
\end{figure}

We provide Figure \ref{fig:Error} of an erroneous tool call to demonstrate several approaches to error handling, where specific responses will be referenced by the codes assigned in the figure (e.g., \ref{fig:Error}.UI to refer to "Order me an onion."). The system prompt, containing role and tool information, is excluded for simplicity.

\subsubsection{Static Retry}

The simplest approach to handling a problematic output is to retry the LLM call with the same prompt. While other \hyperref[hyperparameters]{hyperparameters} may stay the same, the seed should \textit{always} change between static retries to avoid completely duplicate calls. If using a low temperature or a high top-p, then it may also make sense to adjust those values appropriately so as to receive a different output. In the context of Figure \ref{fig:Error}, this might look like \ref{fig:Error}.UI simply being rerun with a different seed.

For low-context calls that yield output that is easily verifiable (e.g., parsing the output into a JSON object), it is a simple yet valuable addition to attempt a few static retries in case verification fails. For outputs that are more difficult to verify, such as natural language instructions that are interpreted downstream, static retries are less helpful as the cost of verification increases.

\subsubsection{Informed Retry} \label{informed-retry}

A more informed approach is to append the LLM’s output to the history, add another user message indicating that the output was unsuccessful, and try again. This should be supplemented with specific error messages or additional directions~\cite{kamoi2024evaluatingllmsdetectingerrors, tyen2024llmsreasoningerrorscorrect}. In the Figure \ref{fig:Error} example, an informed retry might look like sending the following list of messages: \ref{fig:Error}.UI, \ref{fig:Error}.AO, \ref{fig:Error}.TR, and "Attempting the above code yielded the provided error. Please provide an updated output that achieves the initial instruction.". 

\subsubsection{External Retry} \label{external-retry}

Rather than asking for an informed retry from the same context, we can pull out pieces of the history and provide it to an LLM in a separate context to either fix the previous output or generate a new one. This will likely require significant context from the original call but can be supplemented and differentiated by using a different role, different instructions, and error information. Often, shifting roles from, for example, a software engineer to a code reviewer can provide the impetus the LLM needs to fix or generate the correct output. While it has been shown that the explanations from external LLM-based error systems are frequently unreliable and sensitive to prompt changes~\cite{kamoi2024evaluatingllmsdetectingerrors}, having access to task-specific roles, detailed error information (e.g., the error raised by a piece of generated Python code), and background context helps mitigate those issues\cite{tyen2024llmsreasoningerrorscorrect}.

In the Figure \ref{fig:Error} example, an external retry might look like sending the following list of messages: "You are an expert debugger. You have access to \{tool information\}." and "When attempting to fulfill the request, '\{\ref{fig:Error}.UI\}', a helper tried to run the code `\{\ref{fig:Error}.AO\}`, which yielded `\{\ref{fig:Error}.TR\}`. Please provide an updated output that achieves the initial instruction.".

It should be noted that LLMs struggle to locate errors~\cite{kamoi2024evaluatingllmsdetectingerrors, tyen2024llmsreasoningerrorscorrect} but demonstrate strong error correction capabilities if provided sufficient context, specifically error location~\cite{tyen2024llmsreasoningerrorscorrect}. As such, when employing \nameref{informed-retry} or \nameref{external-retry}, care should be taken to include error information that pinpoints the source, such as diagnostic error messages and tracebacks from APIs and runtime environments.

\subsection{Stopping} \label{stopping}

As the control flow of the agentic LLM system is controlled by an LLM, a clear stopping method needs to be defined. This will likely take the form of a predetermined stop token or phrase inserted into the system prompt, such as "TERMINATE"~\cite{wu2024autogen}. It should be a token or phrase that is easily parsable and not otherwise likely, to avoid accidental stopping.

\subsection{Multiple Personas} \label{multiple-personas}

Often, the role that an LLM is assigned has a significant impact on its performance on a given task. This has been observed in LLM literature generally, becoming a key ingredient of effective prompting~\cite{karmaker-santu-feng-2023-teler, kong-etal-2024-better}, and in recent LLM multi-agent research, emerging as a necessary component for agent multiplicity in many such architectures~\cite{Hong2023MetaGPTMP, wu2024autogen, Li2023MetaAgentsSI, Guo2024LargeLM, wang-etal-2024-unleashing, GenAgentsSimulcraPark2023}. For example, while the Figure \ref{fig:Example}.Pe1 role is good for answering most of the user's queries, the Figure \ref{fig:Example}.Pe2 role may be better at answering Figure \ref{fig:Example}.I4 because it requires specialist culinary knowledge.

Because there are likely to be many distinct tasks that form part of an agentic LLM system, there is usually room for multiple roles to be used. An overview of approaches to defining personas for LLMs, or "profiling" them, is detailed in ~\cite{Wang_2024}, categorizing them as handcrafted (e.g.,~\cite{qian-etal-2024-chatdev, GenAgentsSimulcraPark2023}), LLM-generated (e.g,~\cite{wang-etal-2024-unleashing, xu2023expertpromptinginstructinglargelanguage}), or dataset-aligned (i.e., derived from a pertinent dataset). The roles should be informed by the task that the call is handling. This is dependent on the overall context of the agentic LLM system but can largely be addressed in the following ways:

\begin{itemize}
    \item If the tasks are well-defined, handcraft specialist roles for each task (e.g., Figure \ref{fig:Example}.Pe1 and \ref{fig:Example}.Pe2).
    \item If the tasks are not well-defined but generally correspond to a single topic, use the most specific handcrafted role for that topic (e.g., the catch-all Figure \ref{fig:Example}.Pe1).
    \item If the tasks are truly undefined to start (e.g., an assistant that helps with anything) or the topic is very broad:
    \begin{itemize}
        \item Define several distinct roles to which the LLM agent can route subsequent calls as it sees fit~\cite{si2023getting}. Once the agent is in use, a more informed set of personas can be defined according to the most frequently ones. This may also be thought of as the dataset alignment approach~\cite{Wang_2024}, where the dataset is constructed in the environment under an interim set of personas.
        \item Leverage an LLM to create the role that it deems would be best able to respond to the prompt~\cite{wang-etal-2024-unleashing, xu2023expertpromptinginstructinglargelanguage}. This is more expensive as generating the role requires LLM usage but is certainly more robust to unforeseen scenarios. This approach may be used in conjunction with the above point (e.g., if no suitable predefined role is found, create one).
    \end{itemize}
\end{itemize}

\subsection{Managing Relevant Context}

Managing the context that is sent to an LLM is an effective method of increasing the efficiency (speed and cost) and performance of an LLM system, as inference time is dependent on the number of input tokens~\cite{AttIsAllYouNeed, Pope2022EfficientlyST} and LLMs perform worse in long-context scenarios, particularly for complex tasks~\cite{StruggleLongContextLi, liu-etal-2024-lost}. Additionally, careful context management is a necessity given that LLMs have limited context windows\footnote{E.g., Claude 3: 200k~\cite{AnthropicModel}; Gemini 1.5 Pro: 2M+~\cite{GeminiModel}; GPT-4o: 128k~\cite{OAIModel}}. Even for "long-"context LLMs (>100k token limit), many tasks quickly become unwieldy if not properly managed (e.g., working with HTML, where single webpages can be hundreds of thousands of tokens). This is a key consideration to make during \hyperref[task-decomp]{task decomposition}; the more specific the task, the more extraneous context (e.g., prior messages) can be trimmed~\cite{qian2024longllmsnecessitylongcontexttasks}. As such, the context that a specific LLM call receives should be tailored to the task as much as possible. Even if an LLM call requires past messages, it is often possible to strip out certain pieces of context or summarize them, leaving the parts the subsequent call relies on intact and maintaining the overall meaning. Significant adjustments can be made to the context between calls to decrease the overall token count and remove extraneous context, thus reducing LLM confusion and increasing performance for the LLM call~\cite{qian2024longllmsnecessitylongcontexttasks}.

\section{Additional Considerations}

\subsection{Model Size}

The size of the model to use is typically driven by three concerns: cost, speed, and performance. Usually, the bigger the model, the higher the cost, the lower the speed, and the better the performance (although this is not a hard-and-fast rule). It can be tempting to build an agentic LLM system around the weakest model that will adequately do the job so that all three conditions are optimized from the start. However, attempting to build out a functional system from a smaller model first will likely be more time consuming and expensive than starting at the strongest model possible and downgrading the models used for specific calls once the LLM agent has demonstrated competence in the environment. Due to the influence one call can have on subsequent ones, it is infeasible to understand what is possible for a given use case if not all the pieces are working optimally. By starting with stronger models, there will be a gold-standard baseline to compare against so the performance impact of downgrading a model for a specific call can be measured\footnote{See~\cite{TOCost, AWSBig} for discussions on these points from an industry perspective.}. It is recommended that the correct model is selected on a per-task basis and evaluated both individually and in the context of the entire agentic LLM system.

\subsection{Evaluation} \label{eval}

Evaluating an agentic LLM system can be challenging due to the potential for long sequences, non-determinism in LLMs, interactions with external entities, and tasks that may not have obviously correct solutions. Nonetheless, it is essential to have an approach to evaluation defined before deployment to (1) have a baseline to compare against and (2) measure performance changes over time and in response to changes.

When creating a dataset for evaluating an LLM agent, the most important consideration is that it accurately resembles the environment in which is will be deployed. There are many LLM agent benchmarks available targeting specific domains~\cite{deng-etal-2024-mobile, WebshopYao2022, liu2024agentbench, zhang-etal-2024-codeagent} as well as general purpose application~\cite{srivastava2023beyond, wu2024smartplaybenchmarkllmsintelligent, mialon2024gaia}, but many agentic LLM systems applied to a specific task will be too niche to benefit from a broader benchmark. However, insomuch as an established benchmark fits the application of the LLM system, it can be a strong starting point for evaluation and refinement. Whether an existing benchmark is used or not, it is advisable to collect informative agent interactions (e.g., long sequences, short sequences, incorrect outputs, correct outputs, etc.) and related metadata (e.g., hyperparameters) in the deployment environment. Doing so will allow the creation of a dataset, comprised of reproducible input and output pairs, that is derived from the environment. Even a dataset with a few samples will provide a baseline to compare against to ensure prompt engineering addresses failed executions, identify the effects of model and prompt changes, and avoid regression in the system\footnote{See~\cite{LangchainEval} for an industry approach to evaluating deployed LLM systems.}.

While traditional metrics (e.g., precision, recall, etc.) are useful to track, metrics specific to the agent can help reveal changes in the system that higher-level metrics fail to reflect~\cite{EvalSurveyChang2024, kapoor2024aiagentsmatter}. For example, an LLM agent that arrives at the same answer when presented with two different prompts is superficially consistent but a difference in the number of intermediate steps to reach that conclusion may indicate that the system is overly sensitive to prompt changes. Building from~\cite{liu2023llmpempoweringlargelanguage, kapoor2024aiagentsmatter, mehta-etal-2024-improving} that suggest types of alternative evaluation, we provide sample metrics below to use as a starting place, although useful metrics should be chosen in accordance with the design of the LLM agent and the environment in which it is implemented\footnote{Note that the following are focused primarily on evaluating agentic LLM systems but that external components should also be evaluated, such as the RAG system (e.g., the quality of retrievals and the fidelity of embedded documents)~\cite{Salemi2024EvaluatingRQ, es-etal-2024-ragas} and tools (e.g., reliability and consistency of their output).}.

\subsubsection{Holistic} \label{holistic}

No matter how well an agentic LLM system might do along the way or what emergent capabilities it might demonstrate, the final output will determine whether the system is accomplishing its task or not. It is impossible to tell how a composition of LLM calls will perform without running them end-to-end; thus, evaluating an LLM agent should primarily rely on holistic metrics to determine if it is performing as expected.

\textbf{Sample Metrics.}

\begin{itemize}
    \item Across X distinct prompts, how many correct answers does the agent produce?
    \item For input X across N trials, how many distinct answers does the agent produce?
    \item For input X across N trials, what is the average number of steps executed by the agent?
    \item For input X across N trials, what is the average number of tools used by the agent?
    \item For input X (that requires LLM planning) across N trials, what is the average number of steps in each plan?
    \item For input X across N trials, what is the average cost/time?
\end{itemize}

\subsubsection{Piecemeal} \label{piecemeal}

Measuring the performance of a single or a subset of LLM calls that completes a definable task is a viable method of diagnosing problems in or making changes to the system. However, due to the influence a single LLM call can have downstream in an LLM agent, isolated piecemeal evaluation of an agentic LLM system should never be considered a substitute for \nameref{holistic} measures.

\textbf{Sample Metrics.}

\begin{itemize}
    \item For call X with N trials, how many distinct answers are produced?
    \item For N synonymous versions of input A to call X, how many distinct top-K documents are provided by RAG from each embedded version of A?
    \item For call X with tool access across N trials, how many distinct tools are used?
    \item For call X across N trials, what is the average cost/time?
\end{itemize}

\subsection{Integration with Traditional Engineering} \label{trad-eng}

Because LLMs are inherently stochastic, it is often easier to offload as much of the agent's responsibility onto traditional engineering as possible. This allows outsourcing parts of the system that require determinism to methods that can be deterministic. By crafting an LLM agent according to software engineering best practices, we can ensure that key components that are necessary for a given task are always completed or included, rather than relying on the agent to make a request or execute an action. This can take the form of automatically managing context between calls, \hyperref[io-processing]{output processing}, combining tools into toolsets (e.g., putting Figure \ref{fig:Example}.T2 and \ref{fig:Example}.T3 behind a "delivery" interface), incorporating information from long-term memory permanently into the prompts (e.g., Figure \ref{fig:Example}.E1), setting callbacks on certain transitions and calls (e.g., to generate a summary of the most recent conversation to use as context when transitioning from Figure \ref{fig:Example}.Pe1 to \ref{fig:Example}.Pe2), and adding an evaluation after each step of a plan (see \nameref{p-adherence}). (The last two can be thought of as implicit tool usage; see Section \ref{using-tools}). However, care should be taken not to limit the autonomy of the agent in doing so. One way to return autonomy to the agent while still leveraging the benefits of traditional engineering is to allow the agent to short-circuit.

\subsubsection{Short-Circuiting}

Short-circuiting (from the world of software engineering: the idea of evaluating an expression only so far as to guarantee a single answer) is an integral technique for agentic LLM systems. This can be as simple as including \hyperref[stopping]{stopping} criteria into the LLM agent's instructions (see Section \ref{stopping} for examples) or allowing the LLM agent to produce a final output in a single turn. If an agentic LLM system does not short-circuit when it obviously should, the system may have an overreliance on external engineering (i.e., the flow (or parts of the flow) of the agent being hard-coded)\footnote{A recent example of this is OpenAI’s GPT-o1~\cite{GPTo1}. The initial implementation has no short-circuiting, meaning even simple queries that a much weaker model can handle or that require no significant output still incur a full traversal of the agentic LLM system. For example, asking GPT-o1 to “Do nothing” will still pass through the planning, thinking, and alignment stages of the system.}.

As an example, the query presented in Figure \ref{fig:Example}.I3 demonstrates an instance when an LLM agent may want to short-circuit. The query poses a simple question-answering scenario that most current models could satisfactorily respond to. Allowing the agent the autonomy to determine what step to take next (as opposed to, for example, implicitly calling Figure \ref{fig:Example}.P1 for every input) would permit it to simply provide an answer, thus short-circuiting any other components.

\section{Limitations}

Although we present some practical methods for the evaluation of deployed systems, we do not explore human-in-the-loop evaluation as human-computer interaction represents a rich field of study that exceeds the scope of this work.

An important follow-up to evaluation is \textit{how} to compare and respond to changes in a deployed agentic LLM system, such as prompt, model, and environment changes. These considerations remain largely underexplored in current literature and represent some of the key challenges to deploying real-world LLM agents. We do not discuss these considerations as agent maintenance does not fall into the scope of this work but suggest that they are prominent directions for future work.

We explore one aspect of cost for agentic LLM systems, model size, but leave other considerations (such as whether to use an out-of-the-box model or to finetune one on a specific task~\cite{bucher2024finetunedsmallllmsstill, lehman2023needclinicallanguagemodels}, to leverage increasingly strong open-source models~\cite{dubey2024llama3herdmodels, jiang2023mistral7b, bai2023qwentechnicalreport} or to rely on aligned commercial models~\cite{openai2024gpt4technicalreport, geminiteam2024geminifamilyhighlycapable, claude}, and, similarly, to self-host or to use a 3rd party provider) for future work as cost and feasibility of proposed agent architectures warrant a review on their own. See~\cite{kapoor2024aiagentsmatter} for a discussion on the need for cost-informed LLM agent research.

While we approach the agent's underlying LLM from a black-box perspective for simplicity and relevance to many industry applications, approaching it as whitebox opens up additional complexities and opportunities. We deem that considering model specifics exceeds the scope of this review but recognize the value of future work highlighting practical considerations for the real-world deployment of whitebox LLM agents.

\section{Conclusion}

 In this review, we present relevant research into LLM agents and derive actionable insights from it that can be utilized when implementing and deploying agentic LLM systems in the real world. We ascribe relevant research and insights to the four main components of LLM agents from application-focused literature—\nameref{planning}, \nameref{memory}, \nameref{tools}, and \nameref{control-flow}—to provide a review that is mutually accessible to both industry and academia. Namely, for \nameref{planning}, we explore how poor LLM planning capabilities hinder current LLM agent applications and the practical benefits to be derived from task decomposition; for \nameref{memory}, we explore the benefits of and practical considerations to make when leveraging RAG and long-term memory in an LLM agent; for \nameref{tools}, we discuss how to present and manage tools for an LLM agent; for \nameref{control-flow}, we provide practical insights for promoting an uninterrupted LLM agent execution and managing agent internals, such as personas and context usage; and, lastly, suggest additional considerations, such as model size, evaluation, and integrating an LLM agent with traditional engineering.

\begin{acks}
We would like to acknowledge Sergei Petrov and Sonny George, whose input and feedback were instrumental in shaping the foundations of this work.
\end{acks}

\bibliographystyle{ACM-Reference-Format} 
\bibliography{citations}

\end{document}